\documentclass[10pt]{article}
\usepackage[a4paper,textwidth=15.5cm,textheight=23.4cm,headsep=18pt,footskip=26pt]{geometry}
\usepackage[T1]{fontenc}
\usepackage[utf8]{inputenc}
\usepackage{newpxtext,newpxmath}
\usepackage[scaled=0.88]{helvet}
\usepackage{amsmath}
\usepackage{graphicx}
\usepackage{booktabs}
\usepackage{array}
\usepackage{multirow}
\usepackage{microtype}
\usepackage{xcolor}
\usepackage[explicit]{titlesec}
\usepackage{fancyhdr}
\usepackage[font=small,labelfont={sf,bf},labelsep=space,justification=justified,singlelinecheck=false,skip=5pt]{caption}
\usepackage[round,authoryear,sort]{natbib}
\usepackage[section]{placeins}
\usepackage{enumitem}
\usepackage{xurl}
\usepackage{hyperref}

\definecolor{accent}{HTML}{8E2620}
\definecolor{muted}{HTML}{5F5F5F}
\hypersetup{colorlinks=true,linkcolor=accent,citecolor=accent,urlcolor=accent,pdftitle={Embodiment-aware control by inference over the operator: a simulation study},pdfauthor={Sara Falcone}}
\DeclareCaptionLabelFormat{bar}{#1 #2\,\textcolor{muted}{|}}
\titleformat{\section}{\sffamily\bfseries\large}{\textcolor{accent}{\thesection}}{0.7em}{#1}
\titleformat{\subsection}{\sffamily\bfseries\normalsize}{\textcolor{accent}{\thesubsection}}{0.7em}{#1}
\titleformat{\paragraph}[runin]{\sffamily\bfseries\small}{}{0pt}{#1.}
\titlespacing*{\section}{0pt}{16pt plus 3pt minus 2pt}{6pt}
\titlespacing*{\subsection}{0pt}{11pt plus 2pt minus 2pt}{4pt}
\titlespacing*{\paragraph}{0pt}{6pt plus 1pt}{0.6em}

\fancypagestyle{first}{\fancyhf{}\fancyfoot[C]{\sffamily\footnotesize\thepage}}

\newcommand{\dz}{d_z}
\begin{document}
\thispagestyle{first}

\begin{flushleft}
{\sffamily\small\textcolor{accent}{\textbf{PREPRINT}}\hspace{0.6em}\textcolor{muted}{Version 2, November 2026}}\\[6pt]
{\sffamily\footnotesize\textcolor{muted}{This work has been submitted to the IEEE for possible publication. Copyright may be transferred without notice, after which this version may no longer be accessible.}}\\[10pt]
{\sffamily\bfseries\LARGE Embodiment-aware control by inference\\[3pt] over the operator: a simulation study}\\[12pt]
{\large Sara Falcone}\\[3pt]
{\small\textcolor{muted}{Department of Computer Science, Seidenberg School of Computer Science and Information Systems,\\ Pace University, New York, USA \quad·\quad \href{mailto:sfalcone@pace.edu}{\texttt{sfalcone@pace.edu}}}}
\end{flushleft}
\vspace{4pt}
\noindent{\color{muted}\rule{\textwidth}{0.4pt}}
\vspace{2pt}

\noindent{\sffamily\bfseries\small Abstract}\quad
{\small Teleoperation and assistive devices are tuned for channel fidelity, while whether the operator experiences the device as part of the body, the Sense of Embodiment, is measured only afterwards. Minimizing the mismatch between the operator's predictions and the returned feedback fails twice: the predictions cannot be observed, and removing feedback minimizes the objective. We formulate an embodiment-aware controller that instead infers the operator's embodiment and visuo-proprioceptive cue weighting from implicit gaze and pupil signals and task outcome, and chooses bounded device settings under explicit preferences, cast as a discrete Active Inference agent. In simulations with heterogeneous synthetic operators, it raised embodiment from 0.76 to 1.66 on a 0--2 scale over the best fixed setting, within 0.02 of an oracle across ten random seeds; adaptation that did not read the operator gained nothing, and a prediction-error minimizer withheld feedback. An expected-utility controller with the same inference did as well, so the benefit comes from inference over the operator with explicit preferences, not from information seeking. Rules on a single signal came close when that signal was informative and failed when it was not; the controller degraded less, but failed when it trusted an uninformative signal unless briefly calibrated. Against published human data, the model reproduced the embodiment-performance relation, and embodiment developed on a plausible time scale. The embodiment signals were redundant when task success tracked embodiment and became the controller's objective when it did not. Studies with people must first establish calibrated cue-weighting signals and effort-independent embodiment signals.}

\vspace{6pt}

\vspace{2pt}
\noindent{\color{muted}\rule{\textwidth}{0.4pt}}

\section{Introduction}
When a person operates a device that mediates action, such as a teleoperated arm, a robotic orthosis or a prosthesis, the nervous system must come to treat an engineered body, the surrogate, as its own. This Sense of Embodiment (SoE), usually described through the senses of ownership, agency and self-location \citep{kilteni2012,longo2008,falcone2023}, supports how users learn with a device \citep{falcone2024} and is implicated in whether they keep using it \citep{makin2017}, at a time when a substantial fraction of upper-limb prosthesis users still abandon their device \citep{biddiss2007,salminger2022}. Engineering practice optimizes something else. Teleoperation and assistive devices are tuned for mechanical transparency, the fidelity with which motion and force cross the channel \citep{lawrence1993}. More fidelity is not always better for the operator: rendering an environment's mass and friction made teleoperated tasks slower and more forceful \citep{vanderwalt2025}, and feedback may need to be only as faithful as the operator can perceive and use \citep{hirche2012,vanderwalt2024}. Prosthetics now engineers embodiment through sensory feedback \citep{marasco2018}, but these designs are open-loop: they do not measure, while the device is in use, whether embodiment has been achieved. Embodiment is still assessed afterwards by questionnaire, and questionnaire reports are themselves sensitive to expectation and suggestion \citep{lush2020}.

Predictive-processing and Active Inference accounts offer a way to make embodiment a control variable. On these accounts the body is the set of sensory channels whose inputs are best predicted by the agent's own generative model, and embodiment corresponds to low, precision-weighted prediction error across visual, proprioceptive and tactile channels \citep{friston2010,apps2014,limanowski2013}. Building on the proposal that embodiment of a telerobot yields perceptual transparency \citep{toet2020}, a design goal follows, which we call Predictive Transparency: adapt the device so that what it returns matches what the operator predicts. Two obstacles stand in the way. First, the operator's predictions are not observable, so a controller cannot compute the operator's prediction error. Second, an objective that only penalizes mismatch has a trivial minimizer, removing the feedback that carries the mismatch, and it asks for more than embodiment needs: embodiment tolerates moderate inconsistencies between the device and the operator's body \citep{toet2020}.

This paper addresses both obstacles with one formulation. We treat the controller, the Universal Embodiment Engine (UEE), as an Active Inference agent \citep{parr2022,dacosta2020} whose world is the operator. The UEE does not compute the operator's prediction error. It maintains a generative model of the operator in which embodiment and cue weighting are hidden states, observed through candidate implicit signatures (pupil diameter, gaze transition entropy, gaze--hand lead time and look-ahead-time ratio; \citealp{falcone2022b,krejtz2015,land2001}) together with task outcome. It acts on bounded device settings by minimizing expected free energy under explicit preferences for task success and for high-embodiment signatures, and two safeguards, a feedback-gain floor and a minimum dwell between changes, constrain what it may do. This is the logic of inferring another agent's hidden state \citep{friston2015duet}, applied to a human--machine interface. Active Inference is one formulation of Bayesian control over such a hidden state: it makes the preferences explicit and adds an information-seeking term to the choice of action. To find out which of these ingredients matter, we compare the UEE with an expected-utility controller that uses the same inference without the information-seeking term, with model-free bandits that adapt without a model of the operator, and with threshold rules of the kind used in adaptive automation.

We ask four questions in simulation, with heterogeneous synthetic operators whose parameters the controller does not know: (i) can the UEE infer an operator's cue weighting from implicit signals within seconds; (ii) does controlling on that inference raise embodiment and performance relative to fixed, fidelity-optimized control, to adaptation that does not read the operator, to model-free adaptation, and to rules on single signals; (iii) does the naive formulation, minimizing a prediction-error signal, collapse as expected, and which ingredient of the UEE prevents it; and (iv) how does the benefit degrade when the operators depart from the controller's model, including operators from outside its model family. The simulation does not validate the signals, which only user studies can do; it tests the controller's logic and identifies the conditions under which it helps or fails.

The contributions are: (1) a formulation of embodiment-aware control as Bayesian inference over the operator's hidden state, cast as Active Inference, in which the controller infers the operator's embodiment and cue weighting instead of computing the operator's prediction error; (2) a demonstration that a naive prediction-error objective collapses to withholding feedback, with an ablation of the preferences and safeguards that prevent it; (3) a simulation showing that inference over the operator, not adaptation as such and not information seeking, brings embodiment close to an oracle's, and that its advantage over simpler rules lies in robustness when individual signals fail; (4) a demonstration that embodiment signals become necessary, as the controller's objective, when task performance does not track embodiment; and (5) a grounding of the model against published statistics, together with robustness and out-of-family analyses across random seeds, that turn the controller's failure modes into requirements for empirical studies.

\section{Related work}
The UEE draws on several lines of work: how embodiment is measured, how the brain weighs visual and proprioceptive cues, how devices and interfaces adapt to their users, how robots infer a person's hidden state, what the implicit signals it reads are known to measure, how embodiment has been adapted to individuals, and how Active Inference has been used in robotics.

\paragraph{Measuring the Sense of Embodiment} Embodiment of an external object was first shown with the rubber hand illusion, in which synchronous stroking of a seen rubber hand and the hidden real hand makes the rubber hand feel like one's own \citep{botvinick1998}. Its measures fall into three groups. Questionnaires address ownership, agency and self-location, and have been consolidated into a standardized instrument for avatars \citep{gonzalezfranco2018,peck2021}. Behavioural measures include proprioceptive drift, the shift of the felt position of the real hand toward the artificial one \citep{tsakiris2005}, and the cross-modal congruency effect \citep{verhagen2020}. Physiological measures include the skin conductance response to a threat to the artificial limb \citep{armel2003} and pupil dilation \citep{falcone2022b}. Their outcomes depend on methodological choices \citep{riemer2019}, and explicit and implicit measures can diverge: in one study, visuo-proprioceptive synchrony changed ownership ratings and performance but had little effect on proprioceptive measures \citep{falconetaylor2024}; reviews of prosthetic embodiment report the same variety of definitions and measures \citep{zbinden2022}. Almost all of them are taken after a block or in response to a probe such as a threat, which is why embodiment is assessed rather than controlled. The UEE needs signals that are available every second without probes and treats each of them as uncertain evidence, not as a measurement.

\paragraph{Cue weighting} When vision and proprioception disagree about the position of the hand, the brain combines them, weighting each by its reliability \citep{vanbeers1999,ernst2002}. Whether two signals are combined at all depends on whether they are inferred to share a cause \citep{kording2007}, and the rubber hand illusion has been modeled in the same way, as causal inference about whether the seen hand is one's own \citep{samad2015}. Reliability, and with it the weighting, can differ between people, tasks and devices, so a device that delivers corrective feedback through one channel suits some operators better than others. This is the heterogeneity the UEE exploits: it treats the operator's cue weighting as a hidden state to be inferred, not as a population constant.

\paragraph{Adapting devices and interfaces to the user} Human-in-the-loop optimization tunes a device to each user by optimizing a measured objective online, for example the metabolic cost of walking with an exoskeleton or a soft exosuit \citep{zhang2017,ding2018}. These methods treat the user as a black box and need an objective that can be measured directly. Physiological computing and adaptive automation close a similar loop on psychophysiological signals: a biocybernetic loop adjusts a system to the user's inferred state \citep{fairclough2009,ewing2016}, adaptive automation reallocates tasks between operator and machine according to measured engagement or workload \citep{byrne1996,prinzel2000,parasuraman2000}, and passive brain-computer interfaces use implicit brain signals to adapt human-machine systems \citep{zander2011}. These systems usually map a signal to an adaptation through a threshold or a learned classifier, and the state they target is workload or engagement. Embodiment cannot be measured directly, which is why the UEE keeps a model of the operator's hidden state and treats the implicit signals as evidence about it. The bandit baselines in this paper stand for model-free adaptation of this kind.

\paragraph{Inferring the user's hidden state} Shared-autonomy methods infer a user's latent goal from their inputs and act to assist it under uncertainty, either by formulating assistance as a partially observable Markov decision process \citep{javdani2018} or by blending user and robot commands according to the confidence of the inferred goal \citep{dragan2013}; intent detection and arbitration are reviewed in \citet{losey2018}. The hidden state need not be a goal. Robots have planned with a person's trust as a latent state \citep{chen2018}, inferred how willing a user is to adapt to the robot \citep{nikolaidis2017}, and chosen actions for their effect on what people will do \citep{sadigh2016}. The UEE applies the same logic to a different hidden state: not what the operator wants to do, but whether the device is experienced as part of the body and which senses the operator relies on.

\paragraph{Implicit signals of embodiment and effort} The signals the UEE reads are established measures of effort and visuomotor control. Pupil size grows with processing load \citep{beatty1982}, and pupil-based indices of cognitive load are used in human-computer interaction \citep{duchowski2018}. Gaze transition entropy indexes how structured visual scanning is \citep{krejtz2015,shiferaw2019}. In manipulation, gaze leads the hand to the landmarks of the task \citep{land2001,johansson2001}, and becomes predictive as a new visuomotor mapping is learned \citep{sailer2005}. Their link to embodiment is more recent and partly indirect: the pupil responds to threats to an embodied surrogate and, during tasks, may follow embodiment through workload \citep{falcone2022b}. This dual role is why the UEE treats them as evidence about embodiment and cue weighting rather than as measures of either, and why the grounding analysis asks which of them the benefit depends on.

\paragraph{Adaptive embodiment} In virtual reality, avatar properties have been adapted to individuals, for example by using reinforcement learning to find how much movement distortion a person tolerates before embodiment breaks, with physiological signals proposed as future inputs \citep{porssut2022}. Predictive-processing accounts of embodiment \citep{apps2014,limanowski2013} have been brought to teleoperation \citep{toet2020}, but not, as far as we know, as the controller of the interface.

\paragraph{Active Inference in robotics} Active Inference has been applied to robot control and state estimation, for example to reaching with a robot arm \citep{piolopez2016}, to body estimation and adaptation through predictive coding \citep{lanillos2018}, and to body perception and action on a humanoid robot \citep{oliver2022}; its use in robotics is reviewed in \citet{lanillos2021}. In this work the agent's generative model is of the robot's own body and environment. The UEE turns this around: the generative model is of the operator, and the controlled variables are the device settings that shape the operator's experience. It thereby combines the strands above: online, model-based inference of embodiment and cue weighting from implicit signals, used to choose device settings.

\section{The UEE as Active Inference over the operator}
\label{sec:model}
\begin{figure}[!tb]
\centering
\includegraphics[width=\textwidth]{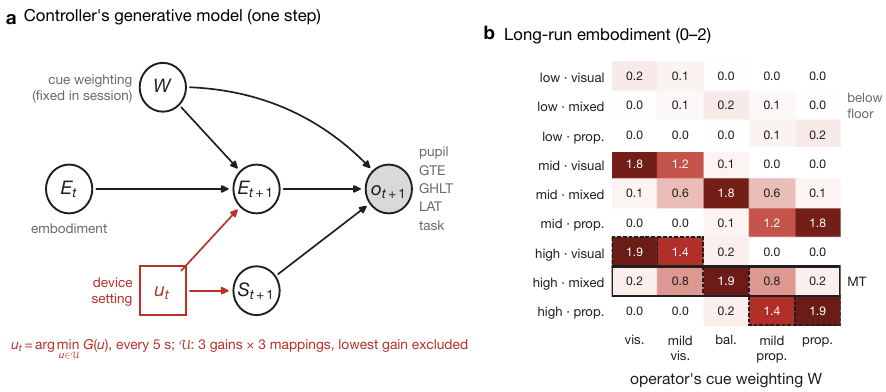}
\caption{\textbf{Generative model and problem structure.} \textbf{a},~One step of the controller's generative model. The operator's embodiment $E$ and cue weighting $W$ are hidden (open circles); the five observation modalities $o$ are observed (grey). The controller's action $u_t$ sets the device, $S_{t+1}=u_t$, and changes the operator's embodiment dynamics. \textbf{b},~Long-run mean embodiment level of an operator held at each setting (rows) for each cue-weighting type (columns), under the population model. Dashed cells mark the best setting for each type (the Oracle); the outlined row is the best single setting for the population (Mechanical Transparency, MT). The two extreme types need opposite mappings, and the two mild types benefit from the extreme mapping on their side but less than the extreme types do.}
\label{fig:model}
\end{figure}

Figure~\ref{fig:model}a shows one step of the controller's generative model. Time advances in steps of one second.

\paragraph{Hidden states} The model has three state factors: the operator's embodiment level $E$ (low, mid or high); the operator's visuo-proprioceptive cue weighting $W$, five types located at $\omega_W \in \{-1, -0.6, 0, +0.6, +1\}$ on an axis from vision-dominant to proprioception-dominant (vision-dominant, mildly vision, balanced, mildly proprioceptive, proprioception-dominant); and the device setting $S$, one of nine combinations of feedback gain $g$ (low, mid or high) and visuo-motor mapping $m$ (visual, mixed or proprioceptive, located at $\omega_m \in \{-1, 0, +1\}$). The mapping stands for any device parameter whose benefit depends on which sense the operator relies on, for example whether corrective feedback is delivered through the visual scene, the haptic channel or both. We describe the settings in teleoperation terms, but they have direct counterparts in other devices: in a prosthesis, the gain could be the intensity of vibrotactile or electrotactile feedback and the mapping whether feedback reaches the user through vision or through the residual limb.

\paragraph{Dynamics} The action $u_t$ takes effect immediately: $S_{t+1} = u_t$, and embodiment evolves under the new setting, $E_{t+1} \sim B(E_{t+1} \mid E_t, W, u_t)$. Cue weighting is stable within a session, $W_{t+1} = W_t$ (a drift variant is examined in Section~\ref{sec:robust}). Embodiment rises when the setting suits the operator and delivers enough sensory cues, and falls otherwise. The match between mapping and operator is $\mu(W, m) = \exp[-(\omega_W - \omega_m)^2 / 2\sigma_\mu^2]$ with $\sigma_\mu = 0.5$, and the cue factor of the three gains is $\kappa(g) = (0.2, 0.7, 1.0)$. Embodiment moves up one level with probability $p_\uparrow = r_\uparrow\,\mu\,\kappa(g)$ and down one level with probability $p_\downarrow = r_\downarrow\,(1-\mu) + \lambda\,[g=\text{low}] + \ell$ (no move beyond the ends of the scale), with $r_\uparrow = r_\downarrow = \lambda = 0.25$ and a lapse rate $\ell = 0.03$ per second in the population model. The term $\lambda$ encodes the observation that removing cues lowers embodiment, in line with the additive contribution of perceptual cues to the SoE \citep{falcone2022a}. The lapse term lets embodiment drop even under a suitable setting, so no level is absorbing and the long-run embodiment level varies continuously with the match (Figure~\ref{fig:model}b).

\paragraph{Observations} Five modalities are observed each second. Each implicit signal has a latent mean $\nu \in [0,1]$ that is mapped to three ordered bins centred at 0, 0.5 and 1 by a Gaussian kernel ($p(k) \propto \exp[-(c_k-\nu)^2 / 2\sigma^2]$, $\sigma = 0.55$, and $\sigma = 0.35$ for gaze--hand lead time), then mixed with a uniform distribution by a noise level $\varepsilon$ ($\varepsilon = 0.1$ in the controller's model). With $e = E/2$, the latent means are: pupil diameter $0.75 - 0.5\,e + 0.4\,c(W,S)$, where $c = \kappa(g)\,(1-\mu)$ is the sensory conflict delivered by the setting; gaze transition entropy (GTE) $1 - e$; gaze--hand lead time (GHLT) $0.5 + 0.35\,\omega_W + 0.2\,(e - 0.5)$; and look-ahead-time ratio (LAT) $e$, each clipped to $[0,1]$. Task outcome (success or error) is not mixed with noise; success has probability $(0.45, 0.70, 0.85)_g + (-0.1, 0, 0.1)_E$. The pupil therefore responds to both low embodiment and conflict, GHLT is the modality that carries direct information about cue weighting, and task success depends on embodiment as well as on gain. With three GHLT bins and five types, neighbouring types are hard to tell apart from GHLT alone. Section~\ref{sec:robust} relaxes these assumptions.

\paragraph{Inference and action} Beliefs $q(E, W, S)$ are updated each second by fixed-point iteration of a mean-field variational posterior, as implemented in \texttt{pymdp} \citep{heins2022}. For each admissible setting $u$ the controller evaluates the expected free energy of the next step,
\begin{equation}
G(u) = \sum_{j} \Big( \underbrace{D_{\mathrm{KL}}\!\left[q(o_j \mid u)\,\|\,\tilde p(o_j)\right]}_{\text{risk}} + \underbrace{\mathbb{E}_{q(E,W,S \mid u)}\!\left[\mathrm{H}[p(o_j \mid E,W,S)]\right]}_{\text{ambiguity}} \Big),
\label{eq:efe}
\end{equation}
summed over modalities $j$, where $q(E,W,S \mid u)$ is the belief propagated one step under $u$ and $\tilde p(o_j) = \sigma(C_j)$ encodes preferences \citep{dacosta2020,parr2022}. The risk term contains the (negative) predictive entropy, so Eq.~\eqref{eq:efe} equals negative expected utility minus expected information gain about the states, which is how \texttt{pymdp} computes it. Policies are one step deep, and the controller chooses $\arg\min_u G(u)$; exact ties, which occur when two settings are equivalent under the current belief, are broken in favour of keeping the current setting. The likelihoods and transitions are fixed at the population values and are not learned during a session. Preferences favour task success ($C = \pm 2$) and the signatures of high embodiment, low GTE and high LAT ($C = (+1, 0, -1)$ and $(-1, 0, +1)$ over the three bins); no preference is placed on pupil diameter or GHLT. Two safeguards complete the controller: a feedback-gain floor, which removes the three lowest-gain settings from the admissible set, and a dwell of five seconds, so that a new setting is chosen at most every fifth step while beliefs continue to update every second. The prior over embodiment is $(0.5, 0.35, 0.15)$, the prior over cue weighting is uniform, and sessions start at the Mechanical Transparency setting (high gain, mixed mapping).

\section{Simulation design}
\paragraph{Synthetic operators} Each synthetic operator shares the structure of the controller's model but has individual parameters unknown to the controller: a cue-weighting type drawn uniformly from the five; up and down rates, $r_\uparrow$ and $r_\downarrow$, each scaled by an independent factor drawn from $U(0.7, 1.3)$; a lapse rate scaled by a factor from $U(0.5, 1.5)$; and a signal noise level $\varepsilon$ drawn from $U(0.05, 0.25)$. Each operator starts at the lowest embodiment level.

\paragraph{Conditions} Each operator was run in eleven conditions with the same random stream, so that comparisons are paired. \emph{Mechanical Transparency (MT)} is the fixed high-gain setting with the highest long-run embodiment averaged over the five types, which is the mixed mapping (Figure~\ref{fig:model}b); it stands for fidelity-optimized control tuned for the population. \emph{Yoked} replays the sequence of settings the UEE chose for a different operator, matching the amount and timing of change without reading this operator. The \emph{UEE} is run with and without the dwell. The \emph{naive prediction-error (PE) minimizer} uses the same inference but prefers only small pupil responses ($C = \pm 2$), as a proxy for low prediction error, with no task or embodiment preference and no floor. The \emph{expected-utility controller} is the UEE without the information-gain term: the same inference and preferences, choosing the setting with the highest expected utility. Two \emph{bandits} adapt without a model of the operator: each admissible setting is an arm, a new arm is chosen every five seconds, and the reward per second is task success plus the two high-embodiment signatures (low GTE, high LAT), scaled to $[0,1]$. \emph{Thompson sampling} keeps a Beta posterior on each arm's reward; \emph{UCB} uses the UCB1 rule with an exploration constant of 0.1, the best of four values tried on these operators, which favours the baseline. Two \emph{rules} stand for adaptive automation, which maps measured signals to adaptations \citep{byrne1996,prinzel2000}, and use the same per-second signals. The \emph{threshold rule} maps the mean GHLT bin since the start of the session to the visual mapping below 0.8, the proprioceptive mapping above 1.2 and the mixed mapping otherwise, at high gain, every five seconds. The \emph{trigger rule} keeps the current mapping at high gain and moves to the next one whenever the share of seconds with high-embodiment signatures (low GTE, high LAT) over the last 20~s falls below 0.2. Both were tuned on these operators, the best of 36 and 30 combinations of thresholds and windows, which again favours the baselines. The \emph{Oracle} is fixed at the setting with the highest long-run embodiment for the operator's true type (dashed cells in Figure~\ref{fig:model}b).

\paragraph{Outcomes and analysis} Sessions lasted 120 steps. The primary outcome was the mean embodiment level over the last 60 steps. Secondary outcomes were the mean embodiment level over the whole session, which includes the time needed to identify the operator; the task-success rate; the number of setting changes; the fraction of time at the lowest gain; the time from which the mapping stayed at the one that suits the operator; and the posterior probability of, and maximum-a-posteriori (MAP) accuracy for, the operator's true type. We simulated 300 operators (57 vision-dominant, 49 mildly vision, 54 balanced, 67 mildly proprioceptive, 73 proprioception-dominant) and compared conditions by the mean paired difference, with a normal-approximation 95\% confidence interval, and the standardized paired difference $\dz$ (mean difference divided by its standard deviation). We do not report $p$-values, which simulated samples can make arbitrarily small. The main experiment and the ablation used seed 2026, the robustness analyses seed 7, the out-of-family analyses seed 11 and the grounding analyses seed 13; with deterministic tie-breaking all results are exactly reproducible. To check that the results do not depend on the random stream, the main comparisons were repeated with ten seeds of 300 operators and the key robustness scenarios with ten seeds of 200 operators; we report the mean and standard deviation across seeds.

\paragraph{Parameter grounding} The parameter values in Section~\ref{sec:model} were set by hand. Raw data from the author's earlier studies are not available, so we compared quantities that the model implies with statistics published from those studies and with the literature (Table~\ref{tab:grounding}). Three comparisons agree. Across people, the correlation between mean embodiment and task success, in a simulated within-subject study in which each operator spent 60~s at each of the nine settings, was 0.44 (median 0.39 in simulated samples of 28, interquartile range 0.29--0.49), against 0.36 for ownership and 0.42 for agency in a 28-participant study that crossed five perceptual cues \citep{falcone2022a}. At a suitable setting, embodiment reached the high level after a median of 8~s (interquartile range 5--14~s), a time scale comparable to that on which the rubber hand illusion is reported to develop under synchronous stimulation \citep{ehrsson2004}. The two gain levels bracket the published weight of feedback relative to visuo-proprioceptive match: in the cue study, suppressing tactile feedback lowered ownership by 0.68 and agency by 0.30 of the drop caused by breaking visuo-proprioceptive synchrony (ratio of the mixed-model estimates), and in the model, lowering the gain from high to mid lowered long-run embodiment by 0.09, and from high to low by 0.91, of the drop caused by the least suitable mapping. Two assumptions are not settled by the data. The first is contradicted: the cue study found no interaction between cues, whereas the model combines match and gain multiplicatively, which produces an interaction as large as the main effects. The second, the strength of the implicit signals, is not identified. The pupil moves in the assumed direction in the author's study: pupils were smaller in the condition with higher embodiment during teleoperation \citep{falcone2022b}. This is a session-level effect, however, and does not identify how much each second of signal reveals, and the gaze signals have not yet been related to embodiment. Section~\ref{sec:robust} tests both departures. Our studies report group effects only and do not give individual cue weights \citep{falconetaylor2024}, so the distribution of cue weighting in the population remains an assumption; Section~\ref{sec:robust} replaces it with continuous cue weighting. The preferences, safeguards and priors are design choices of the controller, examined in the ablation.

\begin{table}[t]
\centering
\caption{\textbf{Parameter grounding} against published statistics. Model values from \texttt{grounding.py}.}
\label{tab:grounding}
\small
\begin{tabular}{@{}>{\raggedright\arraybackslash}p{3.6cm}>{\raggedright\arraybackslash}p{3.6cm}>{\raggedright\arraybackslash}p{3.4cm}>{\raggedright\arraybackslash}p{3.6cm}@{}}
\toprule
Quantity & Published & Model & Status \\
\midrule
Embodiment--performance correlation across people & $r = 0.36$--$0.42$, $N = 28$ \citep{falcone2022a} & 0.44; median 0.39 at $N = 28$ & Consistent \\
\addlinespace
Time to high embodiment & seconds (rubber hand illusion) \citep{ehrsson2004} & median 8~s (IQR 5--14) & Plausible \\
\addlinespace
Feedback effect relative to visuo-proprioceptive effect & 0.30--0.68 \citep{falcone2022a} & 0.09 (mid gain), 0.91 (low gain) & Bracketed \\
\addlinespace
Interaction between cues & none \citep{falcone2022a} & as large as the main effects & Not supported; additive cues tested \\
\addlinespace
Implicit signals & pupil: direction \citep{falcone2022b}; gaze: not reported & same direction & Strength not identified; weaker signals tested \\
\addlinespace
Distribution of cue weighting & not reported & five types, uniform & Assumption; continuous weighting tested \\
\addlinespace
Preferences, safeguards, priors & -- & -- & Design choices; ablation \\
\bottomrule
\end{tabular}
\end{table}

\paragraph{What the simulation can and cannot show} The synthetic operators come from the controller's own model family, and in the main population fixed control suits only balanced operators. How close the UEE comes to the Oracle, and how far it rises above fixed control, are therefore partly built into the design. We test this in Section~\ref{sec:robust} with operators from outside the model family (continuous cue weighting and match functions the controller does not assume) and with a sweep over how varied the population is, recomputing the best fixed setting for each population.

\section{Results}
\begin{figure}[!tb]
\centering
\includegraphics[width=\textwidth]{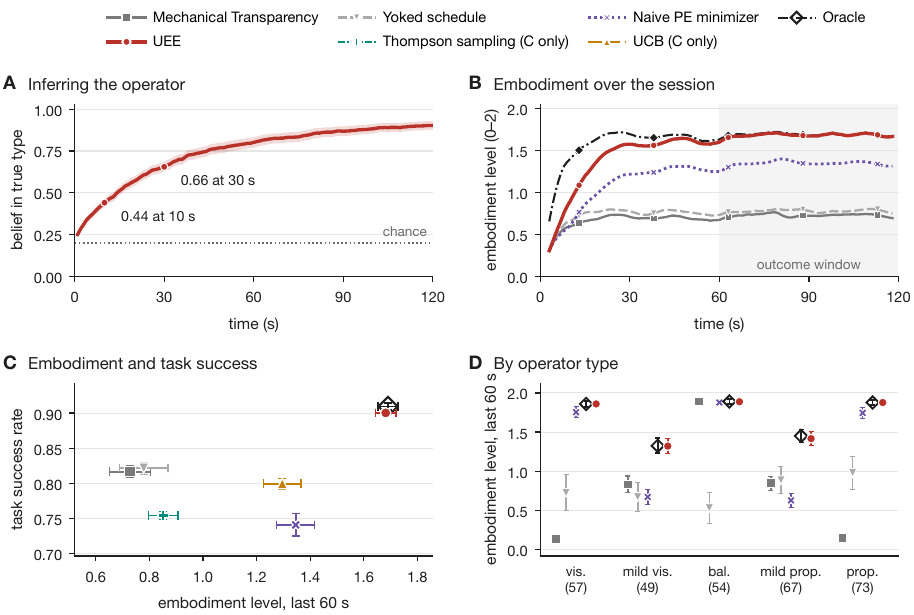}
\caption{\textbf{Main results for 300 synthetic operators.} \textbf{A},~Posterior probability of the operator's true cue-weighting type under the UEE (mean and 95\% CI; chance is 1/5). \textbf{B},~Embodiment level over the session (mean across operators, 5-s moving average); the shaded region is the window of the primary outcome. \textbf{C},~Embodiment in the last 60~s and task-success rate (means and 95\% CIs), including the two model-free bandits; the expected-utility controller coincides with the UEE and is not shown. \textbf{D},~Embodiment in the last 60~s by the operator's type (number of operators in brackets). Mechanical Transparency serves balanced operators well and the others poorly; the UEE comes close to the Oracle for every type.}
\label{fig:results}
\end{figure}

\paragraph{Finding the setting is easier than identifying the operator} Belief in the operator's true type rose from chance ($1/5$) to 0.44 after 10~s, 0.66 after 30~s and 0.80 after 60~s; the MAP type was correct for 54\%, 74\% and 85\% of operators at these times (Figure~\ref{fig:results}A). At 30~s, 78 of the 79 errors confused neighbouring types, and 57 of them confused types that call for the same setting. The decision-relevant question was settled sooner: the mapping reached and kept the one that suits the operator within 30~s for 89\% of operators, with a median of 5~s for the two extreme types, 10~s for balanced operators and 15~s for the two mild types. One mildly proprioceptive operator never settled on the suitable mapping.

\paragraph{Reading the operator, not changing settings, raises embodiment} Table~\ref{tab:main} and Figure~\ref{fig:results}B--C summarize the conditions. In the last 60~s the UEE reached a mean embodiment level of 1.68 (0--2 scale), against 0.73 under Mechanical Transparency (difference 0.95, 95\% CI [0.87, 1.04], $\dz = 1.30$) and 0.78 under the yoked schedule (difference 0.90 [0.81, 1.00], $\dz = 1.07$), and task success rose accordingly (0.90 vs 0.82 and 0.82; $\dz = 1.14$ and $1.01$). The yoked schedule did no better than fixed control (difference 0.05 [$-0.07$, 0.18]), so the benefit comes from reading the operator rather than from changing settings. The benefit was largest for the extreme types, whom the mixed mapping does not suit (Figure~\ref{fig:results}D; 1.86 vs 0.13 and 1.88 vs 0.15), about half as large for the mild types (1.32 vs 0.83 and 1.42 vs 0.85), and absent for balanced operators, for whom MT is already the best setting (1.89 vs 1.89). These results do not depend on the random stream: across ten seeds the UEE's benefit over MT was $+0.90$ (SD 0.03 across seeds; range $+0.86$ to $+0.95$) and over the yoked schedule $+0.93$ (SD 0.05).

\begin{table}[t]
\centering
\caption{\textbf{Outcomes by condition} (300 synthetic operators, 120-s sessions; mean $\pm$ SD).}
\label{tab:main}
\small
\begin{tabular}{@{}lcccc@{}}
\toprule
Condition & Embodiment, last 60 s & Embodiment, session & Task success & Setting changes \\
\midrule
Mechanical Transparency & $0.73 \pm 0.68$ & $0.69 \pm 0.63$ & $0.82 \pm 0.07$ & 0 \\
Yoked schedule          & $0.78 \pm 0.81$ & $0.75 \pm 0.69$ & $0.82 \pm 0.08$ & $1.5 \pm 1.0$ \\
Naive PE minimizer      & $1.35 \pm 0.63$ & $1.19 \pm 0.53$ & $0.74 \pm 0.14$ & $9.0 \pm 3.9$ \\
UEE                     & $1.68 \pm 0.34$ & $1.51 \pm 0.31$ & $0.90 \pm 0.04$ & $1.5 \pm 1.0$ \\
UEE without dwell       & $1.68 \pm 0.35$ & $1.53 \pm 0.32$ & $0.90 \pm 0.04$ & $3.4 \pm 2.7$ \\
Expected-utility controller & $1.68 \pm 0.34$ & $1.51 \pm 0.31$ & $0.90 \pm 0.04$ & $1.5 \pm 1.0$ \\
Thompson sampling       & $0.85 \pm 0.49$ & $0.73 \pm 0.32$ & $0.75 \pm 0.05$ & $18.5 \pm 2.4$ \\
UCB                     & $1.30 \pm 0.62$ & $0.99 \pm 0.43$ & $0.80 \pm 0.07$ & $10.3 \pm 3.9$ \\
Threshold rule (GHLT)   & $1.63 \pm 0.40$ & $1.45 \pm 0.36$ & $0.89 \pm 0.04$ & $1.8 \pm 1.6$ \\
Trigger rule            & $1.51 \pm 0.54$ & $1.22 \pm 0.43$ & $0.87 \pm 0.05$ & $1.4 \pm 1.1$ \\
Oracle                  & $1.69 \pm 0.33$ & $1.61 \pm 0.29$ & $0.91 \pm 0.04$ & 0 \\
\bottomrule
\end{tabular}
\end{table}

\paragraph{The cost of not knowing the operator} In the last 60~s the UEE came within 0.01 of the Oracle (difference $-0.009$ [$-0.016$, $-0.002$], $\dz = -0.14$); its choices differed from the Oracle's in that window for 11 of 300 operators, 7 of them mildly proprioceptive. Over the whole session the gap was ten times larger ($-0.10$ [$-0.12$, $-0.09$], $\dz = -0.81$), and task success was lower by 0.009 ($\dz = -0.66$): 80\% of the cost of identification was paid in the first 30~s and 96\% in the first minute. It was largest for mildly proprioceptive operators (whole-session gap 0.14, against 0.07--0.11 for the other types). Across ten seeds the gap in the last 60~s was $-0.013$ (SD 0.004).

\paragraph{Inference over the operator, not information seeking} The expected-utility controller, which uses the same inference and preferences without the information-gain term, did exactly as well as the UEE: embodiment and task success were identical for all 300 operators (Table~\ref{tab:main}), and within 0.002 in every one of the ten seeds. The model-free bandits did markedly worse. UCB, with its exploration constant tuned on these operators, reached 1.30 (UEE minus UCB 0.39 [0.32, 0.45], $\dz = 0.72$) and Thompson sampling 0.85 (0.83 [0.78, 0.89], $\dz = 1.70$), with lower task success (0.80 and 0.75, against 0.90; Figure~\ref{fig:results}C). Without a model of the operator, the bandits had to learn each setting's value from noisy, delayed signals, and kept switching (10.3 and 18.5 changes per session). What carries the benefit is therefore Bayesian inference over the operator's hidden state with explicit preferences; the information-seeking term of Active Inference adds nothing measurable here.

\paragraph{A rule on the right signal comes close} Rules that map one signal to a setting did much better than the bandits. The threshold rule, which reads GHLT, the signal that carries cue weighting in this model, reached 1.63, only 0.05 below the UEE ([0.03, 0.07], $\dz = 0.28$; 0.047 across ten seeds, SD 0.008), with similar task success (0.89). The trigger rule, which reads the embodiment signatures and searches the mappings in turn, reached 1.51 (0.17 below the UEE [0.13, 0.21], $\dz = 0.49$). When its signal is informative and its thresholds are tuned to the population, a simple rule thus captures most of the benefit. What the rules lack is robustness, examined in Section~\ref{sec:robust}: each depends on one signal and on thresholds fitted to one population.

\paragraph{The naive objective collapses; preferences prevent it} The naive PE minimizer spent 19\% of the session at the lowest feedback gain, where conflict, and with it the pupil response, is smallest. Its task success fell to 0.74 (difference to the UEE 0.16, $\dz = 1.31$) and its embodiment level, although above fixed control, stayed below the UEE's (1.35; $\dz = 0.77$). This collapse is expected by construction: in the model the pupil responds to conflict, and conflict grows with gain, so withholding feedback is what the naive objective asks for. The point is to show the failure concretely and to identify what prevents it. The collapse depended on the operator: for the mild types, whose conflict no mapping removes, the naive controller cut the gain (34--38\% of the session at the lowest gain), embodiment stayed at 0.63--0.67 and task success fell to about 0.60. The ablation in Table~\ref{tab:ablation} separates the ingredients. Information seeking alone, without preferences, did not find the suitable setting: with the floor it stayed at the suitable mapping only 55\% of the time (embodiment 0.93), and without the floor it also withheld feedback (26\% of the session at the lowest gain). Either preference alone was indistinguishable from the full UEE: the task preference favours high gain directly and, because task success depends on embodiment, also the suitable mapping; the embodiment preference favours the suitable mapping directly and, because lapses must be made good, also the high gain that speeds recovery. With the UEE's preferences the floor was never needed. With the naive objective the floor prevented the collapse to the lowest gain, but the controller spent half of the last 60~s at the middle gain, where conflict is smaller, and lost task success (0.81 vs 0.90, $\dz = 1.58$). The pupil's conflict response is thus useful for finding the right mapping, but as the only objective it trades performance for a quieter signal.

\begin{table}[t]
\centering
\caption{\textbf{Ablation of preferences and safeguards} (same 300 operators; means over operators). Objective: which terms enter the choice of setting; expected utility drops the information-gain term, and information only drops the preferences, so that only the expected information gain guides the choice. Matched mapping: fraction of the last 60~s at the mapping that suits the operator.}
\label{tab:ablation}
\small
\begin{tabular}{@{}llccccc@{}}
\toprule
 & & Embodiment, & Task & Time at & Mid gain, & Matched \\
Objective & Floor & last 60 s & success & lowest gain & last 60 s & mapping \\
\midrule
UEE (task + embodiment)  & yes & 1.68 & 0.90 & 0\%  & 0\%  & 99\% \\
UEE (task + embodiment)  & no  & 1.68 & 0.90 & 0\%  & 0\%  & 99\% \\
Expected utility only & yes & 1.68 & 0.90 & 0\%  & 0\%  & 99\% \\
Task only                & yes & 1.68 & 0.90 & 0\%  & 0\%  & 99\% \\
Embodiment only          & yes & 1.68 & 0.90 & 0\%  & 0\%  & 99\% \\
Information only         & yes & 0.93 & 0.84 & 0\%  & 0\%  & 55\% \\
Information only         & no  & 0.78 & 0.72 & 26\% & 0\%  & 64\% \\
Naive (small pupil)      & yes & 1.60 & 0.81 & 0\%  & 53\% & 98\% \\
Naive (small pupil)      & no  & 1.35 & 0.74 & 19\% & 34\% & 98\% \\
\bottomrule
\end{tabular}
\end{table}

\paragraph{The dwell costs little} Without the dwell, the UEE changed setting 3.4 times per session instead of 1.5, and slightly raised whole-session embodiment (1.53 vs 1.51; difference 0.02 [0.01, 0.03]); embodiment in the last 60~s was the same (difference 0.00 [$-0.00$, 0.00]). The dwell's intended purpose, keeping the controller slower than the operator's own adaptation to it, cannot be evaluated here, because the synthetic operators do not adapt to the controller.

\paragraph{The design choices do not drive the result} The preferences, safeguards and priors were set by hand. Varying them one at a time (five seeds of 200 operators each) left the benefit over MT between $+0.88$ and $+0.89$, against $+0.88$ for the reference: task and embodiment preferences at half and one and a half times their values, a dwell of 3 or 10~s, a flat prior over embodiment, no gain floor, and an assumed signal noise of 0.05 or 0.20. Only the dwell changed behaviour, from 2.0 setting changes per session at 3~s to 1.1 at 10~s.

\paragraph{Embodiment signals matter when task success does not track embodiment} In the population model task success depends on embodiment, so a preference for task success already favours the suitable mapping, and either preference alone matched the full UEE (Table~\ref{tab:ablation}). We therefore tested a world in which task success carries no information about embodiment (Table~\ref{tab:proxy}; 200 operators, seed 13, controller calibrated to that world). With the embodiment signals, the full UEE and a controller with the embodiment preference alone kept their benefit ($+0.83$ [0.73, 0.93]), whereas a controller with the task preference alone fell to $+0.14$ [0.05, 0.24]: it kept the gain high but had no reason to choose the suitable mapping. Without embodiment signals, task success alone sufficed when it tracked embodiment ($+0.82$), and the embodiment preference alone fell to $+0.27$ [0.17, 0.36]. When neither carried information, every controller fell below fixed control ($-0.24$ [$-0.34$, $-0.15$]): with no preference that tracks embodiment, the information-gain term moved it away from the population's best setting for 30\% of the last 60~s, whereas the expected-utility controller stayed there and lost nothing ($0.00$). The embodiment signals are therefore redundant when task performance tracks embodiment and become the controller's objective when it does not.

\begin{table}[t]
\centering
\caption{\textbf{When task success does not track embodiment} (200 synthetic operators per row, seed 13). UEE minus Mechanical Transparency in embodiment over the last 60~s, for controllers with both preferences, the task preference only or the embodiment preference only.}
\label{tab:proxy}
\small
\begin{tabular}{@{}llccc@{}}
\toprule
Task success tracks embodiment & Embodiment signals & Both & Task only & Embodiment only \\
\midrule
yes & yes & $+0.83$ & $+0.83$ & $+0.83$ \\
yes & no  & $+0.82$ & $+0.82$ & $+0.27$ \\
no  & yes & $+0.83$ & $+0.14$ & $+0.83$ \\
no  & no  & $-0.24$ & $-0.24$ & $-0.24$ \\
\bottomrule
\end{tabular}
\end{table}

\section{Robustness and operators outside the model family}
\label{sec:robust}
\begin{figure}[!tb]
\centering
\includegraphics[width=\textwidth]{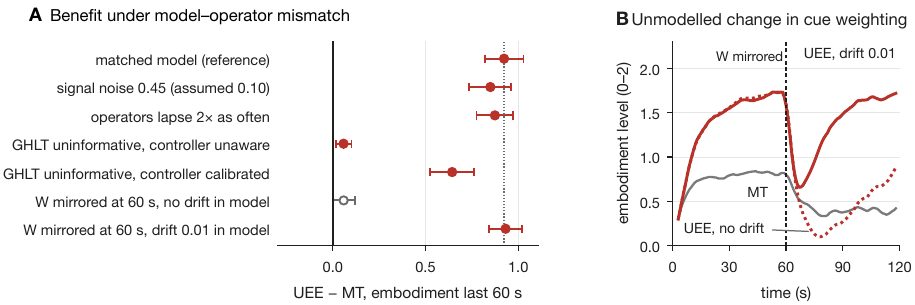}
\caption{\textbf{Where the controller fails.} \textbf{A},~Paired difference in embodiment (last 60~s) between the UEE and Mechanical Transparency, with 95\% CI, for selected scenarios (200 operators each; all scenarios in Table~\ref{tab:robust}). Filled: CI excludes zero; open: it does not. The dotted line marks the reference. \textbf{B},~Embodiment over time (mean across operators, 5-s moving average) when each operator's cue weighting is mirrored at 60~s. A model that assumes stable cue weighting keeps its first inference, falls below fixed control for about 30~s and recovers only partly by the end of the session; a small drift term lets it recover within about 35~s.}
\label{fig:robust}
\end{figure}

\begin{figure}[!tb]
\centering
\includegraphics[width=\textwidth]{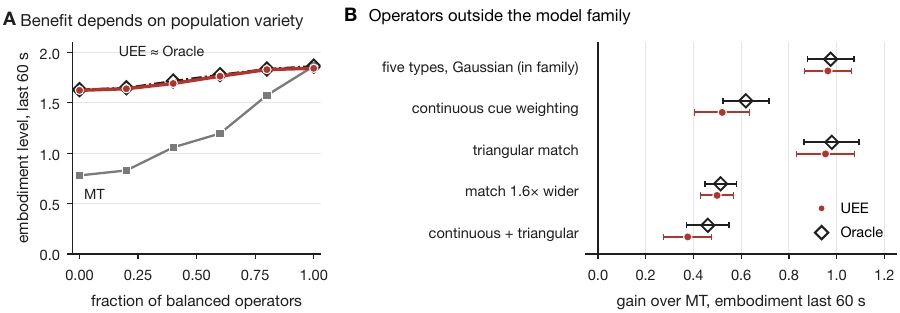}
\caption{\textbf{Population variety and operators outside the model family} (200 operators per point or row). \textbf{A},~Embodiment in the last 60~s as the fraction of balanced operators varies, the rest split evenly over the other four types; Mechanical Transparency is recomputed as the best fixed setting for each population (the visual mapping when there are no balanced operators, the mixed mapping otherwise). \textbf{B},~Gain over Mechanical Transparency (mean and 95\% CI) of the UEE and of the Oracle for operators whose cue weighting is continuous rather than one of five types, or whose match function differs from the one the controller assumes. The controller is unchanged throughout.}
\label{fig:general}
\end{figure}

Each robustness experiment changed one property of 200 new synthetic operators while the controller was unchanged, and compared the UEE with Mechanical Transparency on the same operators (Table~\ref{tab:robust}, Figure~\ref{fig:robust}). With the matched model the UEE's benefit was $+0.92$ [0.82, 1.03]. Two further analyses (Figure~\ref{fig:general}) use operators from outside the controller's model family and vary how heterogeneous the population is.

\begin{table}[t]
\centering
\caption{\textbf{Robustness} (200 synthetic operators per scenario). Difference: UEE minus Mechanical Transparency in embodiment over the last 60~s, with 95\% CI. Belief: posterior probability of the true cue-weighting type at the end of the session.}
\label{tab:robust}
\small
\begin{tabular}{@{}llcc@{}}
\toprule
Property changed & Scenario & Difference [95\% CI] & Belief \\
\midrule
None & matched model (reference) & $+0.92$ [$+0.82$, $+1.03$] & 0.93 \\
\addlinespace
\multirow{4}{*}{Noise (assumed 0.10)} & 0.05 & $+0.93$ [$+0.82$, $+1.03$] & 0.95 \\
 & 0.15 & $+0.92$ [$+0.81$, $+1.02$] & 0.94 \\
 & 0.30 & $+0.89$ [$+0.78$, $+1.00$] & 0.78 \\
 & 0.45 & $+0.85$ [$+0.74$, $+0.96$] & 0.60 \\
\addlinespace
Pupil response to conflict & absent / doubled & $+0.92$ / $+0.92$ & 0.90 / 0.94 \\
Adaptation rate & 2$\times$ slower / 1.6$\times$ faster & $+0.88$ / $+0.91$ & 0.92 / 0.93 \\
Lapse rate & 2$\times$ / 0.5$\times$ & $+0.87$ / $+0.95$ & 0.92 / 0.93 \\
\addlinespace
\multirow{4}{*}{GHLT uninformative} & controller unaware & $+0.06$ [$+0.01$, $+0.10$] & 0.27 \\
 & controller calibrated & $+0.64$ [$+0.53$, $+0.76$] & 0.59 \\
 & \quad + start prior 0.6 & $+0.85$ [$+0.74$, $+0.96$] & 0.85 \\
 & \quad + start prior 0.8 & $+0.89$ [$+0.78$, $+1.00$] & 0.93 \\
\addlinespace
\multirow{4}{*}{$W$ mirrored at 60~s} & no drift in model & $+0.06$ [$-0.00$, $+0.12$] & 0.15 \\
 & drift 0.005 in model & $+0.91$ [$+0.82$, $+1.00$] & 0.71 \\
 & drift 0.010 in model & $+0.93$ [$+0.84$, $+1.02$] & 0.68 \\
 & drift 0.020 in model & $+0.96$ [$+0.88$, $+1.05$] & 0.65 \\
\addlinespace
\multirow{3}{*}{No change: cost of drift} & drift 0.005 & $+0.90$ [$+0.79$, $+1.01$] & 0.83 \\
 & drift 0.010 & $+0.89$ [$+0.79$, $+1.00$] & 0.79 \\
 & drift 0.020 & $+0.86$ [$+0.75$, $+0.97$] & 0.72 \\
\bottomrule
\end{tabular}
\end{table}

\paragraph{Noise, pupil behaviour, adaptation and lapse rates} The benefit degraded gracefully when the signals were noisier than assumed ($+0.85$ at noise 0.45, with final belief in the true type down to 0.60), and changed little when the pupil ignored sensory conflict or over-reacted to it, when operators adapted twice as slowly or 1.6 times as fast as the population model assumed, or when they lapsed twice or half as often. Noise slowed identification; the other properties change how much embodiment a setting yields but not which setting is best, so they barely affected the controller's choices.

\paragraph{A trusted signal must be calibrated} When GHLT carried no information about cue weighting but the controller assumed that it did, the benefit disappeared ($+0.06$ [0.01, 0.10]): uninformative GHLT values looked like evidence for a balanced operator, the controller kept the mixed mapping (90\% of the last 60~s), and belief in the true type stayed near chance (0.27). When the controller's observation model was calibrated to the uninformative signal, about two thirds of the benefit returned ($+0.64$ [0.53, 0.76]). The controller then inferred cue weighting from how the pupil's conflict response, the embodiment signatures and task outcome changed under the settings it chose; the grounding analysis below shows that the pupil's conflict response and task outcome are enough. This is the one scenario in which the information-gain term made a difference, and a small one: without it, embodiment was lower by 0.03 [0.004, 0.048] (23 of 200 operators differed; in the other scenarios we tested, including noise 0.45 and continuous cue weighting, the difference was within $\pm 0.01$). This took longer (median 25--115~s to settle on the suitable mapping for the non-balanced types, against 5--15~s with the matched model) and failed more often for the mild types (suitable mapping during 52\% and 76\% of the last 60~s, against 83--91\% for the other types). A short calibration at the start of the session, represented as a prior of 0.6 or 0.8 on the true type, recovered most of the rest ($+0.85$ and $+0.89$). A calibrated model can partly compensate for the absence of a signal about cue weighting; a miscalibrated one cannot. In this scenario the trigger rule, which does not read GHLT, did better than the unaware or merely calibrated UEE (Table~\ref{tab:seeds}).

\paragraph{Change within a session must be modeled} When each operator's cue weighting was mirrored after 60~s (vision-dominant became proprioception-dominant, mildly vision became mildly proprioceptive, and so on; balanced operators became a random extreme) and the controller assumed that cue weighting could not change, the UEE held on to its first inference (final belief in the new type 0.15). Embodiment fell to 0.10 at 76~s, stayed below that under fixed control for 29~s, and the benefit over the last 60~s was lost ($+0.06$ [$-0.00$, $+0.12$]; Figure~\ref{fig:robust}B). Allowing a small drift in the model's cue weighting (0.005 to 0.02 per second) restored the full benefit ($+0.91$ to $+0.96$), with embodiment back to 90\% of its pre-change level about 35~s after the change. The same drift had a small, growing cost when nothing changed ($+0.90$, $+0.89$ and $+0.86$, against $+0.92$), because a controller that expects change is less certain of what it knows (final belief 0.83 to 0.72).

\paragraph{Operators outside the model family} The controller assumes five cue-weighting types and a Gaussian match function. When operators instead had a continuous cue weighting drawn uniformly from $[-1, 1]$, the UEE's gain over fixed control fell from $+0.96$ [0.86, 1.06] for operators in the family to $+0.52$ [0.41, 0.63], and its gap to the Oracle widened from $-0.01$ to $-0.10$ [$-0.14$, $-0.06$]; it captured 84\% of the Oracle's gain rather than 99\% (Figure~\ref{fig:general}B). Part of the smaller gain is not the controller's doing: with continuous weighting more operators sit near the middle, where fixed control does well, so the Oracle's own gain also fell ($+0.62$). A triangular match function cost little ($+0.95$; 97\% of the Oracle's gain); a match function 1.6 times wider than assumed halved the room for improvement for every controller ($+0.50$, Oracle $+0.51$). Combining continuous weighting with the triangular match gave the smallest gain ($+0.38$ [0.28, 0.48]; 82\% of the Oracle's). The controller's model thus degrades gracefully when it is wrong in form, but the degradation is real and largest when the operator's cue weighting falls between the types the model knows.

\paragraph{The benefit depends on how varied the population is} With no balanced operators, the best fixed setting was the visual mapping and the UEE's gain was $+0.84$ [0.72, 0.96]; the gain fell steadily as the share of balanced operators grew, to $+0.26$ [0.18, 0.34] at 80\%, and was slightly negative when all operators were balanced ($-0.02$ [$-0.04$, $-0.01$]), the cost of an adaptive controller in a population that a fixed setting already serves (Figure~\ref{fig:general}A). At every point the UEE stayed within 0.03 of the Oracle. The size of the benefit over fixed control is therefore a property of the population more than of the controller, and it should be estimated from how much real operators differ.

\paragraph{Each rule fails with its signal} Table~\ref{tab:seeds} repeats the key scenarios with ten seeds, for the UEE and the two rules with their thresholds unchanged. The scenarios are stable across seeds (SD 0.01--0.06). With the matched model the threshold rule came within 0.05 of the UEE. With noisier signals (0.45) it lost a quarter of its benefit ($+0.65$ against $+0.81$ for the UEE), and when GHLT carried no information about cue weighting it lost all of it ($-0.03$); the trigger rule lost everything when the embodiment signatures carried no information ($-0.03$). The UEE combines the signals and degraded less in each case, with one exception: when it trusted an uninformative GHLT it gained only $+0.07$, less than the trigger rule ($+0.71$), and calibrating its model brought it to $+0.63$; only with a short calibration at the start of the session ($+0.85$) did it do better than the rule again. The advantage of inference over a rule on one signal is robustness to the failure of individual signals, and it holds only for a calibrated model.

\begin{table}[t]
\centering
\caption{\textbf{UEE and rules across ten seeds} (200 synthetic operators per seed and scenario). Difference from Mechanical Transparency in embodiment over the last 60~s, mean (SD across seeds). The rules keep the thresholds tuned on the main operators.}
\label{tab:seeds}
\small
\begin{tabular}{@{}lcccc@{}}
\toprule
Scenario & UEE & Threshold rule & Trigger rule & Oracle \\
\midrule
Matched model & $+0.88$ (0.04) & $+0.83$ (0.05) & $+0.71$ (0.05) & $+0.90$ (0.04) \\
Noise 0.45 & $+0.81$ (0.05) & $+0.65$ (0.06) & $+0.53$ (0.05) & $+0.91$ (0.05) \\
GHLT uninformative, controller unaware & $+0.07$ (0.02) & $-0.03$ (0.01) & $+0.71$ (0.05) & $+0.90$ (0.04) \\
\quad controller calibrated & $+0.63$ (0.03) & $-0.03$ (0.01) & $+0.71$ (0.05) & $+0.90$ (0.04) \\
\quad + start prior 0.8 & $+0.85$ (0.05) & $-0.03$ (0.01) & $+0.71$ (0.05) & $+0.90$ (0.04) \\
No embodiment information in the signals & $+0.88$ (0.04) & $+0.87$ (0.05) & $-0.03$ (0.05) & $+0.90$ (0.04) \\
Additive cue combination & $+0.62$ (0.03) & $+0.59$ (0.03) & $+0.35$ (0.03) & $+0.63$ (0.03) \\
\bottomrule
\end{tabular}
\end{table}

\paragraph{Departures suggested by the published data} The grounding (Table~\ref{tab:grounding}) identified two departures, tested on 200 new operators (Table~\ref{tab:grounded}). First, the strength of the implicit signals. Weakening how strongly all four implicit signals depend on embodiment, to 75\%, 50\%, 25\% and 0\% of the population model, while the controller kept assuming full strength, left the benefit unchanged ($+0.83$ [0.73, 0.93] at full strength, $+0.82$ [0.72, 0.92] at zero). In this model, where task success tracks embodiment, the benefit therefore comes from inferring the operator's cue weighting, through gaze--hand lead time and the pupil's response to conflict, not from the embodiment signatures; when task success does not track embodiment, the signatures become the objective (Table~\ref{tab:proxy}). When gaze--hand lead time also carried no information about cue weighting and the controller was calibrated to that, the benefit was $+0.55$ to $+0.66$ across signal strengths; the differences between these values are within the variation between random seeds (the same scenario gave $+0.64$ with seed 7, Table~\ref{tab:robust}), so weaker embodiment signals did not reduce it further, and with no embodiment information at all the controller inferred cue weighting from the pupil's conflict response and task outcome alone ($+0.60$ [0.48, 0.72]). A session-start calibration again recovered the benefit ($+0.79$ and $+0.83$ with a prior of 0.6 and 0.8 on the true type, at half-strength signals). Second, the combination of cues. Operators whose match and gain combine additively, as in the cue study, gained less from adaptation ($+0.56$ [0.48, 0.63]), because fixed control already served them better (1.13 against 0.83), but the UEE, whose model remained multiplicative, still captured 98\% of the Oracle's gain over fixed control, with full or half-strength signals.

\begin{table}[t]
\centering
\caption{\textbf{Departures suggested by the published data} (200 synthetic operators per scenario, seed 13; the controller assumes the population model unless calibrated). Difference: UEE minus Mechanical Transparency in embodiment over the last 60~s, with 95\% CI. Share: fraction of the Oracle's gain over Mechanical Transparency captured by the UEE. Signals: strength of the dependence of the four implicit signals on embodiment.}
\label{tab:grounded}
\small
\begin{tabular}{@{}llccc@{}}
\toprule
Scenario & Signals & Difference [95\% CI] & Share & MT \\
\midrule
Matched model (reference) & 1.00 & $+0.83$ [$+0.73$, $+0.93$] & 99\% & 0.83 \\
\addlinespace
\multirow{4}{*}{Weaker signals, controller unaware} & 0.75 & $+0.83$ [$+0.73$, $+0.93$] & 99\% & 0.83 \\
 & 0.50 & $+0.82$ [$+0.72$, $+0.92$] & 98\% & 0.83 \\
 & 0.25 & $+0.82$ [$+0.72$, $+0.92$] & 98\% & 0.83 \\
 & 0.00 & $+0.82$ [$+0.72$, $+0.92$] & 98\% & 0.83 \\
\addlinespace
\multirow{4}{*}{\shortstack[l]{GHLT uninformative,\\ controller calibrated}} & 1.00 & $+0.55$ [$+0.43$, $+0.67$] & 65\% & 0.83 \\
 & 0.50 & $+0.66$ [$+0.55$, $+0.78$] & 79\% & 0.83 \\
 & 0.25 & $+0.63$ [$+0.51$, $+0.75$] & 75\% & 0.83 \\
 & 0.00 & $+0.60$ [$+0.48$, $+0.72$] & 72\% & 0.83 \\
\addlinespace
\quad + start prior 0.6 / 0.8 & 0.50 & $+0.79$ / $+0.83$ & 94\% / 99\% & 0.83 \\
\addlinespace
\multirow{2}{*}{Additive cue combination} & 1.00 & $+0.56$ [$+0.48$, $+0.63$] & 98\% & 1.13 \\
 & 0.50 & $+0.56$ [$+0.48$, $+0.63$] & 98\% & 1.13 \\
\bottomrule
\end{tabular}
\end{table}

\section{Implications for human-machine system design}
\label{sec:implications}
The simulation tests the logic of an embodiment-aware controller, not real signals, but its results translate into six recommendations for designing such controllers and for the studies that evaluate them.

\paragraph{Put embodiment in the objective when performance does not track it} A controller that optimized task success alone chose the suitable mapping only because task success depended on embodiment; when it did not, that controller kept the feedback high but gained $+0.14$ instead of $+0.83$ (Table~\ref{tab:proxy}). Whether performance tracks embodiment is therefore a property of the population and task that should be measured before a controller is built. Where it does, a performance objective suffices; where it does not, embodiment signatures must enter the objective, and the controller is only as good as their validity.

\paragraph{Find the signal of cue weighting and calibrate it} The benefit came from inferring which senses the operator relies on, and it vanished when the controller trusted a signal that carried no such information (Table~\ref{tab:seeds}). The relation between any candidate signal, such as gaze--hand lead time, and cue weighting should be validated before the signal is used for control, and the controller should include a short calibration at the start of each session. Behavioural cue weights from a brief visuo-proprioceptive conflict task \citep{ernst2002,samad2015} are a natural choice; in the simulation such a calibration restored the benefit even when the online signal was uninformative ($+0.85$). Without it, a calibrated controller still inferred cue weighting from how the pupil's response to conflict and task outcome changed under its own choices, which makes the pupil's sensitivity to visuo-proprioceptive conflict, separately from effort, the next assumption to test.

\paragraph{Choose a model when signals can fail, a rule when one is reliable} A rule tuned on one informative signal came within 0.05 of the UEE, at a fraction of the complexity, and is easier to inspect. It failed completely when its signal failed, and its thresholds are tied to the population it was tuned on. Where one signal has been validated and the population is stable, a rule is a reasonable choice; where signals are uncertain, noisy or may fail, a calibrated model that combines them is the safer one. In both cases the controller should be compared with a rule, not only with fixed control.

\paragraph{Bound what the controller may do} An objective built on a proxy of prediction error alone removed feedback, as its objective implies, and information seeking without a preference that tracks embodiment moved the device away from its best setting ($-0.24$). Explicit preferences for task success or embodiment prevented both; a floor on feedback gain is a cheap guarantee, and a minimum dwell between changes kept the controller to about 1.5 changes per two-minute session at no cost in embodiment. In assistive and teleoperation settings, where an operator must be able to predict the device, these bounds belong in the design rather than in tuning.

\paragraph{Expect the operator to change} A controller that assumed stable cue weighting held on to its first inference when the operator changed, and fell below fixed control; a small drift term restored the benefit at a modest cost. Operators learn, tire and adapt to the device, so the controller should forget at a rate matched to how fast real operators change, which has to be measured.

\paragraph{Design evaluations that can show what the controller adds} Three comparisons separate the contributions of an adaptive interface: fixed control tuned for the population, a yoked schedule that changes settings as often without reading the operator, and a rule on the best single signal. The size of the benefit over fixed control depended on the population, from none when a fixed setting suited everyone to $+0.84$ (Figure~\ref{fig:general}A), so studies should estimate how much their operators differ, for example in behavioural cue weights, rather than assume it. Because pupil size and gaze entropy also respond to effort \citep{beatty1982,shiferaw2019}, workload should be measured and varied independently, so that an apparent embodiment signal is not an effort signal \citep{falcone2022b}. Finally, studies should report whether task performance tracks embodiment in their population, since this decides whether embodiment needs to be in the objective at all.

\section{Discussion}
Treating the controller as an agent that infers the operator, rather than as a device that computes the operator's prediction error, removes the first obstacle to embodiment-aware control. Making preferences explicit removes the second: the naive objective found the trivial solution of withholding feedback, information seeking without preferences gained little (0.93, against 0.73 for fixed control and 1.68 for the UEE), and either a preference for task success or one for high-embodiment signatures prevented the collapse and kept the feedback gain high, with the feedback floor as a guarantee rather than the active ingredient. In simulation the resulting controller came close to an oracle within about a minute, consistently across random seeds, and the yoked comparison showed that the benefit comes from reading the operator, not from adapting as such. A tuned rule on the right signal came within 0.05 of the UEE, however, so a benefit over fixed control alone does not show that a model of the operator is needed.

\paragraph{The role of Active Inference} The comparisons with an expected-utility controller and with bandits separate three ingredients. A model of the operator's hidden state mattered most against model-free adaptation: bandits, which had to learn each setting's value from noisy and delayed signals, fell well short. Against rules that map one signal to a setting, its advantage was smaller in the matched model and lay in robustness: it combined the signals, degraded less with noise and survived the loss of either signal, provided it was calibrated. Explicit preferences mattered next: without them the controller did not find the suitable setting. The information-gain term, the ingredient specific to Active Inference, made no measurable difference except when cue weighting had to be learned from the consequences of the controller's own choices, and there only a small one. The contribution of this paper is therefore Bayesian inference over the operator's hidden state with explicit preferences; Active Inference is the formulation we used, and its value lies in expressing inference, preferences and information seeking in one objective and in connecting the controller to predictive accounts of embodiment, not in a performance advantage in this setting. It can also cost: when no preference tracked embodiment, information seeking alone moved the controller away from the population's best setting and lowered embodiment below fixed control ($-0.24$), which the expected-utility controller avoided. Information seeking may matter more when the signals are weaker, the settings richer, or the horizon longer than one step, which is where the formulation should next be tested.

The controller did not need to know exactly who the operator was. With five cue-weighting types and signals that barely separate neighbouring types, the most probable type was still wrong for a quarter of operators after 30~s, yet the controller had already settled on the suitable setting for most of them. A controller that chooses actions by their expected consequences, with or without the information-gain term, resolves the uncertainty that matters for its next choice and can leave the rest unresolved. The cost of identification was real but concentrated in the first half-minute, and the size of the benefit over fixed control depended on how varied the population was, from $+0.84$ with no balanced operators to none when all were balanced.

The embodiment signals play two roles. As evidence, they added little in this model: the benefit came from inferring cue weighting. As an objective, they were indispensable whenever task success did not track embodiment: a controller that optimized task success alone then kept the feedback high but had no reason to choose the mapping that suits the operator. This is the case an embodiment-aware controller is meant for. In long-term use of a prosthesis or a teleoperated arm, task performance may be maintained by compensation while embodiment is lost; the device would then optimize the wrong quantity unless embodiment itself is in its objective.

The robustness and grounding analyses also state what an empirical program must establish first, in this order: that some signal, or a short calibration, reveals the operator's cue weighting; that the embodiment signatures track embodiment rather than effort; and how fast operators change within a session (Section~\ref{sec:implications}).

\paragraph{Limitations} The main synthetic operators belong to the same model family as the controller, with individual variation. Operators with continuous cue weighting or other match functions reduced the benefit but did not remove it, yet they remain close relatives of the controller's model; the simulation tests the controller's logic, not the relation between real signals and embodiment. States, signals and settings are coarsely discretized; cue weighting is one-dimensional; planning looks one step ahead; and the operator's adaptation to the controller is represented only through the embodiment dynamics, so the value of the dwell could not be assessed. Parameter values were set by hand and checked against published statistics rather than fitted to raw data, which were not available: the check supports the embodiment dynamics and the relation between embodiment and performance, brackets the relative weight of the cues, and leaves the strength of the signals and the population's distribution of cue weighting as assumptions (Table~\ref{tab:grounding}). The model's multiplicative combination of cues is not supported by the data; operators with additive cues reduced the benefit but not the controller's share of the Oracle's gain. The preferences and safeguard settings are design choices, not fitted, although varying them did not change the result. The rule-based baselines were tuned on the main operators, which favours them in the matched model; other rules, for example combining several signals, may do better. The embodiment scale is abstract and not calibrated to questionnaire scores. A physical implementation also needs the mechanical safeguards of bilateral teleoperation, such as passivity-based control, which are independent of the inference.

\paragraph{Conclusion} Embodiment can be formulated as a control variable without observing the operator's predictions, by inferring the operator. In simulation, Bayesian inference over the operator with explicit preferences comes close to oracle performance when its model is right, degrades gracefully when the operators depart from it in form, needs to resolve only the uncertainty that matters for its choices, and fails in informative ways when its model is wrong. It matches a tuned rule when one signal is reliable and is more robust when signals fail, provided it is calibrated, and it needs embodiment signals as its objective whenever performance does not track embodiment. Its failures define the measurements that must come first: calibrated, embodiment-specific signals, and a model of how the operator changes.

\paragraph{Code and data availability}Code, results and figure scripts: \url{https://github.com/sa-falcone/uee-embodiment-sim}, archived at \href{https://doi.org/10.5281/zenodo.24106761}{doi:10.5281/zenodo.24106761} (version 1.1.0); the parameter grounding is in \nolinkurl{grounding.py} and the revision analyses in \nolinkurl{analyses_v2.py} and \nolinkurl{analyses_v2b.py}.
The simulation uses \texttt{inferactively-pymdp} 1.0.4; each analysis runs in minutes on a laptop.

\footnotesize
\bibliographystyle{plainnat}
\setlength{\bibsep}{0.4pt plus 0.3pt}
\bibliography{refs}

\end{document}